\documentclass[runningheads]{llncs}
\usepackage[T1]{fontenc}
\usepackage{graphicx}
\usepackage{algorithmic}
\usepackage{algorithm}
\usepackage{color}
\usepackage{amsmath}

\usepackage{caption}
\usepackage{amsmath}
\usepackage{subfigure}

\usepackage{amsfonts} 
\usepackage{amssymb}

\usepackage{marvosym}

\begin{document}
\title{Enabling DBSCAN for Very Large-Scale High-Dimensional Spaces}

%
\author{Yongyu Wang}
%
%
\institute{JD Logistics, Beijing 101111, China\\
\email{wangyongyu1@jd.com}}

\maketitle              

\renewcommand{\thefootnote}{}
\footnotetext{orcid:0009-0006-0705-752X}

\renewcommand{\thefootnote}{}

\begin{abstract}
DBSCAN is one of the most important non-parametric unsupervised data analysis tools. By applying DBSCAN to a dataset, two key analytical results can be obtained: (1) clustering data points based on density distribution and (2) identifying outliers in the dataset. However, the time complexity of the DBSCAN algorithm is $O(n^2 \beta)$, where $n$ is the number of data points and $\beta = O(D)$, with $D$ representing the dimensionality of the data space. As a result, DBSCAN becomes computationally infeasible when both $n$ and $D$ are large. In this paper, we propose a DBSCAN method based on spectral data compression, capable of efficiently processing datasets with a large number of data points ($n$) and high dimensionality ($D$). By preserving only the most critical structural information during the compression process, our method effectively removes substantial redundancy and noise. Consequently, the solution quality of DBSCAN is significantly improved, enabling more accurate and reliable results.

\keywords{DBSCAN \and Spectral Graph Theory \and Data Compression.}
\end{abstract}
\section{Introduction}
\label{sec:intro}

One of the most fundamental and widely used tasks in data analysis, especially in practical applications, is partitioning data into different clusters and identifying anomalies within the dataset. DBSCAN is an algorithm that uniquely achieves both tasks in a single run, making it highly popular in both academia and industry. Moreover, compared to other mainstream clustering algorithms, such as k-means, spectral clustering, and deep neural network-based clustering methods, DBSCAN does not require prior specification of the number of clusters in the data. This is a critical advantage when working with previously unseen datasets, as it facilitates a more intuitive understanding of the data.

However, DBSCAN also has a notable drawback—its high computational complexity. A bottleneck of the original DBSCAN algorithm lies in the need to perform a range query for each data item, i.e., calculating the number of neighbors within the distance \( \varepsilon \). Consequently, the overall time complexity can reach \( O(n^2 \beta) \) in the worst case, where \( n \) is the number of data items and \( \beta \) represents the complexity of computing the distance between two items. For instance, if the data is a set of points in \( \mathbb{R}^D \), \( \beta = O(D) \). When \( n \) or \( \beta \) is large, the range query process can significantly slow down the execution of DBSCAN. This limitation makes DBSCAN unsuitable for the increasingly common demands of large-scale and high-dimensional data analysis.

To address this issue, researchers have devoted considerable effort to accelerating DBSCAN. However, our experiments on benchmark datasets from the National Institute of Standards and Technology and the UC Irvine Machine Learning Repository reveal that existing methods invariably face a trade-off between speed and accuracy. None of these approaches fully achieve the ultimate goal of significantly accelerating the algorithm while maintaining the quality of DBSCAN's output. 

In this paper, a novel method has been proposed for accelerating DBSCAN based on spectrum-preserving data compression, enabling DBSCAN to operate efficiently in very large-scale high-dimensional spaces. our proposed method not only achieves acceleration without compromising DBSCAN's solution quality but also significantly enhances the quality of its solutions. This approach paves the way for broader applications of this classic algorithm across various fields.

\section{Preliminary}

\subsection{DBSCAN}

Given a dataset \( D \) with \( N \) samples \( \mathbf{x}_1, \dots, \mathbf{x}_N \), the DBSCAN (Density-Based Spatial Clustering of Applications with Noise) algorithm identifies clusters through two main steps: neighborhood search and cluster expansion.

In the \textbf{neighborhood search step}, DBSCAN defines clusters based on the density of data points. For a given point \(\mathbf{x}_i\), the algorithm examines all points within a specified radius \(\varepsilon\) (referred to as the \(\varepsilon\)-neighborhood). A point \(\mathbf{x}_i\) is classified as a \textbf{core point} if the number of points within its \(\varepsilon\)-neighborhood, including \(\mathbf{x}_i\) itself, is at least a minimum threshold \(\text{MinPts}\). Points that do not meet this criterion are classified either as \textbf{border points} (if they lie within the \(\varepsilon\)-neighborhood of a core point) or as \textbf{noise} if they do not belong to any cluster.

Once core points are identified, DBSCAN proceeds to the \textbf{cluster expansion step}. Here, each core point initiates a cluster by linking to other core points within its \(\varepsilon\)-neighborhood, iteratively expanding the cluster by including reachable core points and their neighbors. This expansion continues until all reachable points within the density threshold are included in the cluster. The process is repeated for all core points, eventually identifying all clusters in the dataset.

Formally, DBSCAN can be summarized by the following steps:

\begin{enumerate}
    \item \textbf{Identify Core Points}: For each point \(\mathbf{x}_i\), compute the \(\varepsilon\)-neighborhood and count the number of points within it. If the count meets or exceeds \(\text{MinPts}\), classify \(\mathbf{x}_i\) as a core point.

    \item \textbf{Cluster Expansion}: For each core point, form a new cluster and iteratively add all directly reachable core points within \(\varepsilon\)-neighborhoods. Expand the cluster until no more points meet the density-reachability criteria.

    \item \textbf{Classify Border and Noise Points}: Points within the \(\varepsilon\)-neighborhood of core points but with fewer than \(\text{MinPts}\) neighbors are classified as border points. Points that do not fall within any cluster are classified as noise.
\end{enumerate}

DBSCAN excels in detecting clusters of arbitrary shapes and identifying noise, making it a robust choice for diverse clustering tasks. However, its efficiency can be significantly impacted by computational limitations. Specifically, the neighborhood search step, which requires calculating pairwise distances for each point, can become a bottleneck. This issue is exacerbated in large-scale datasets, where the sheer volume of points amplifies the computational burden, and in high-dimensional spaces, where distance computations are inherently more resource-intensive.

\subsection{Graph Laplacian Matrices}

Consider a graph \( G = (V, E, w) \), where \( V \) represents the set of vertices, \( E \) is the set of edges, and \( w \) is a function assigning positive weights to each edge. The Laplacian matrix of graph \( G \), which is a symmetric diagonally dominant (SDD) matrix, is defined as follows:

\begin{equation}\label{formula_laplacian}
L_G(p,q) = 
\begin{cases}
-w(p,q) & \text{if } (p,q) \in E, \\
\sum\limits_{(p,t) \in E} w(p,t) & \text{if } p = q, \\
0 & \text{otherwise}.
\end{cases}
\end{equation}

\begin{figure}[!h] \centering 
\subfigure[$k$-means on the orig space of the two circles data set]{ \includegraphics[width=1.5in,height=1in]{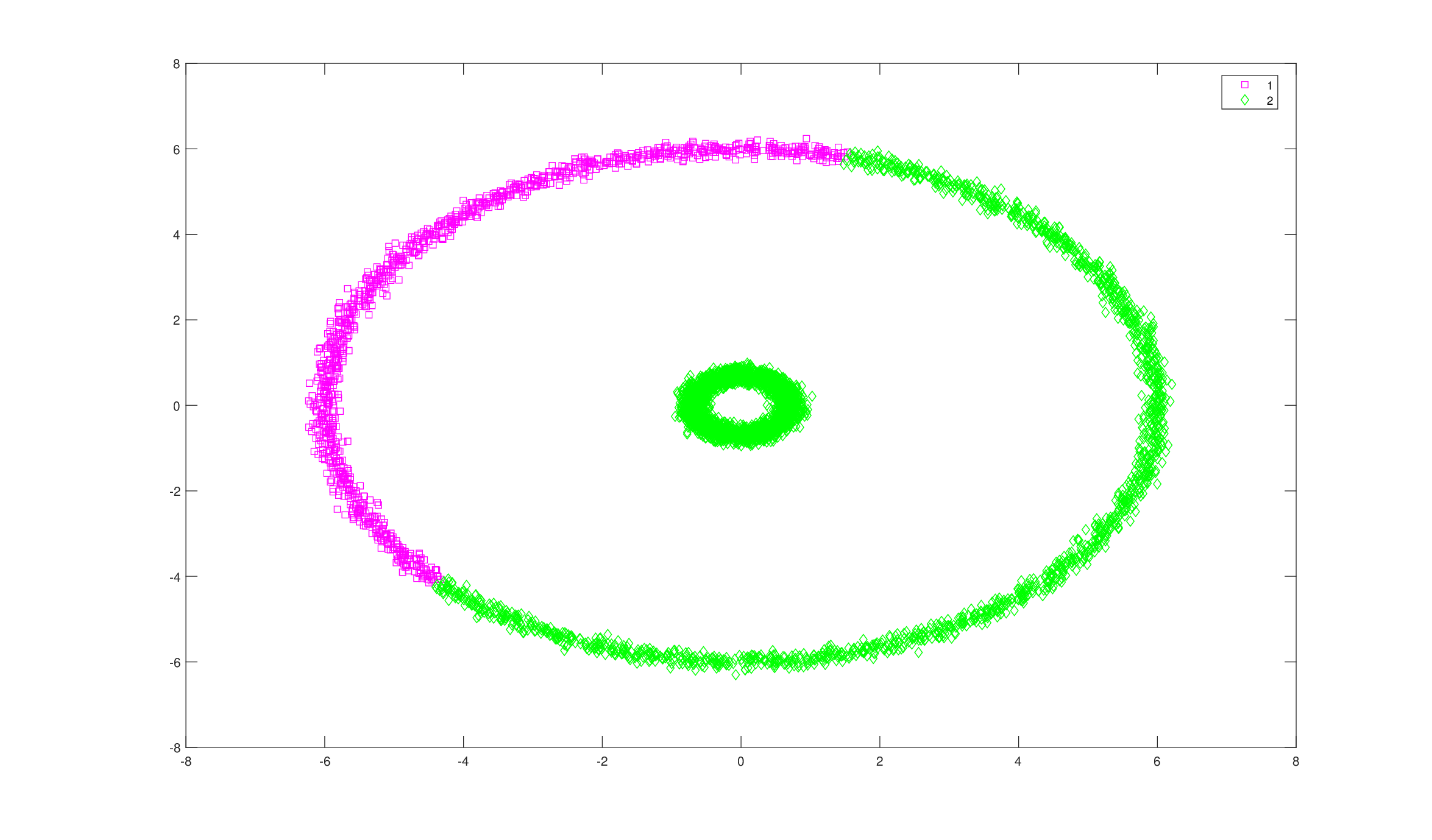} } \subfigure[$k$-means on the spectrally-embedded space of the two circles data set ]{ \includegraphics[width=1.5in,height=1in]{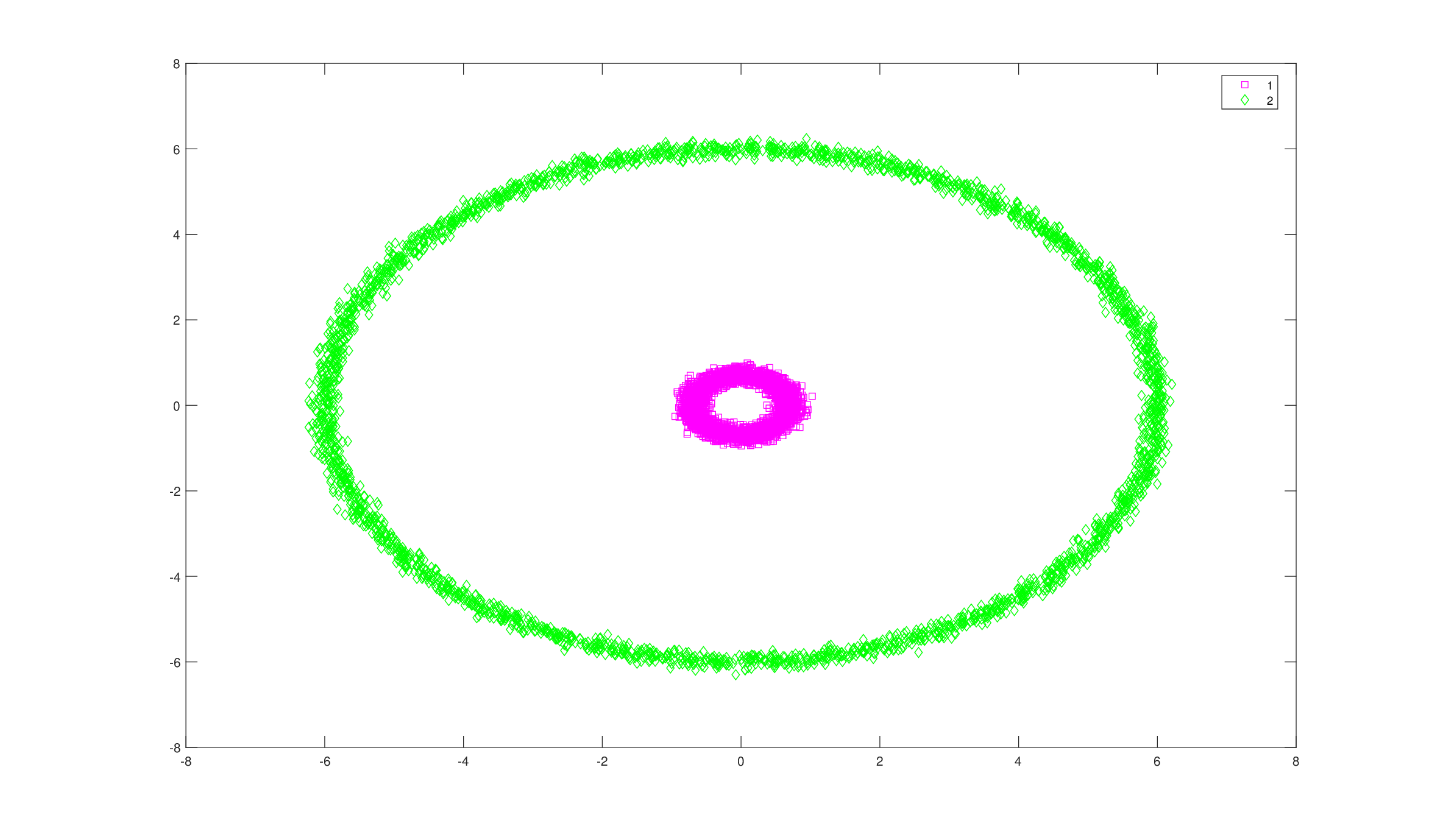} } \subfigure[$k$-means on the orig space the two circles data set]{ \includegraphics[width=1.5in,height=1in]{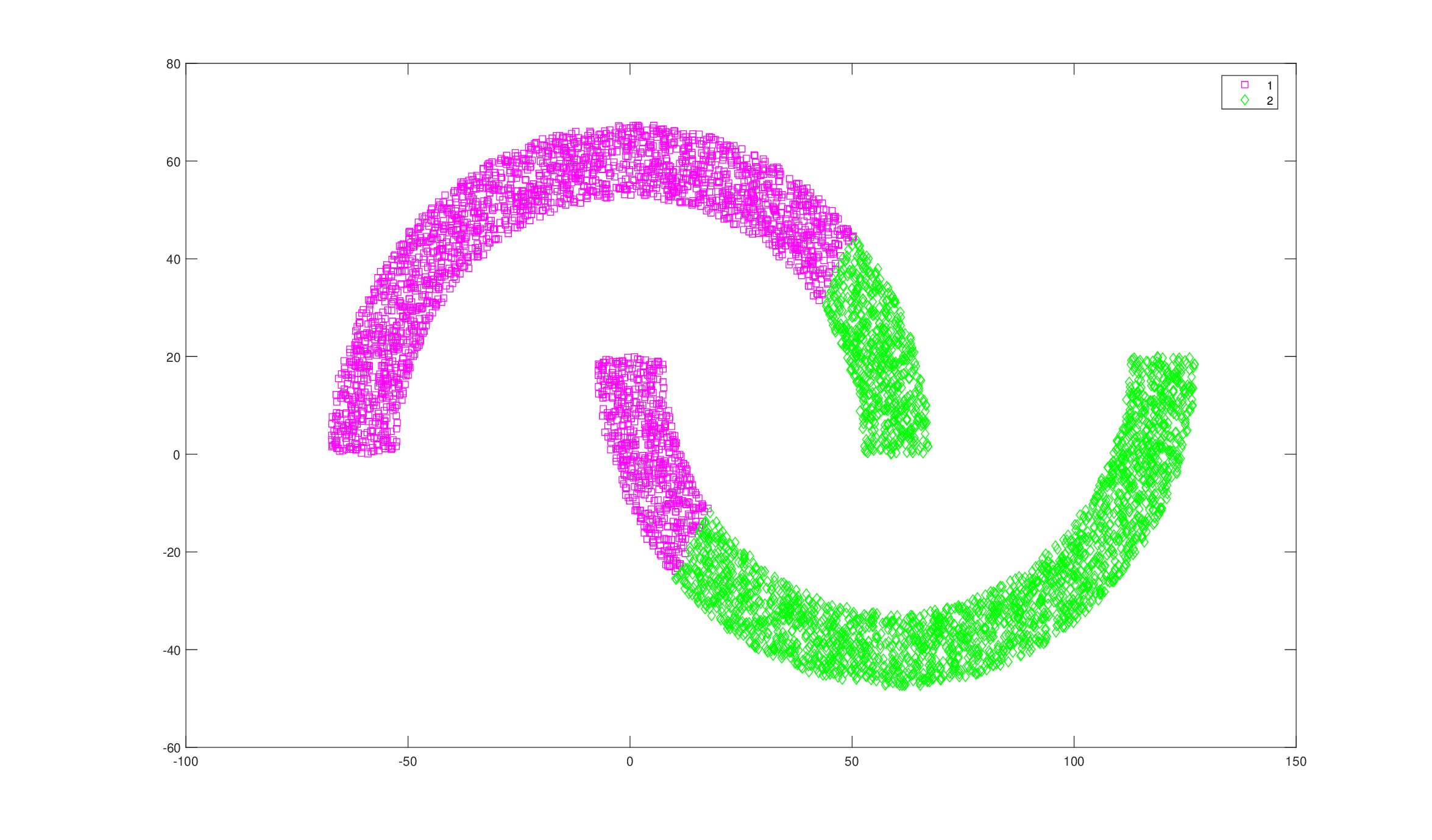} } \subfigure[$k$-means on the spectrally-embedded space of two circles data set]{ \includegraphics[width=1.5in,height=1in]{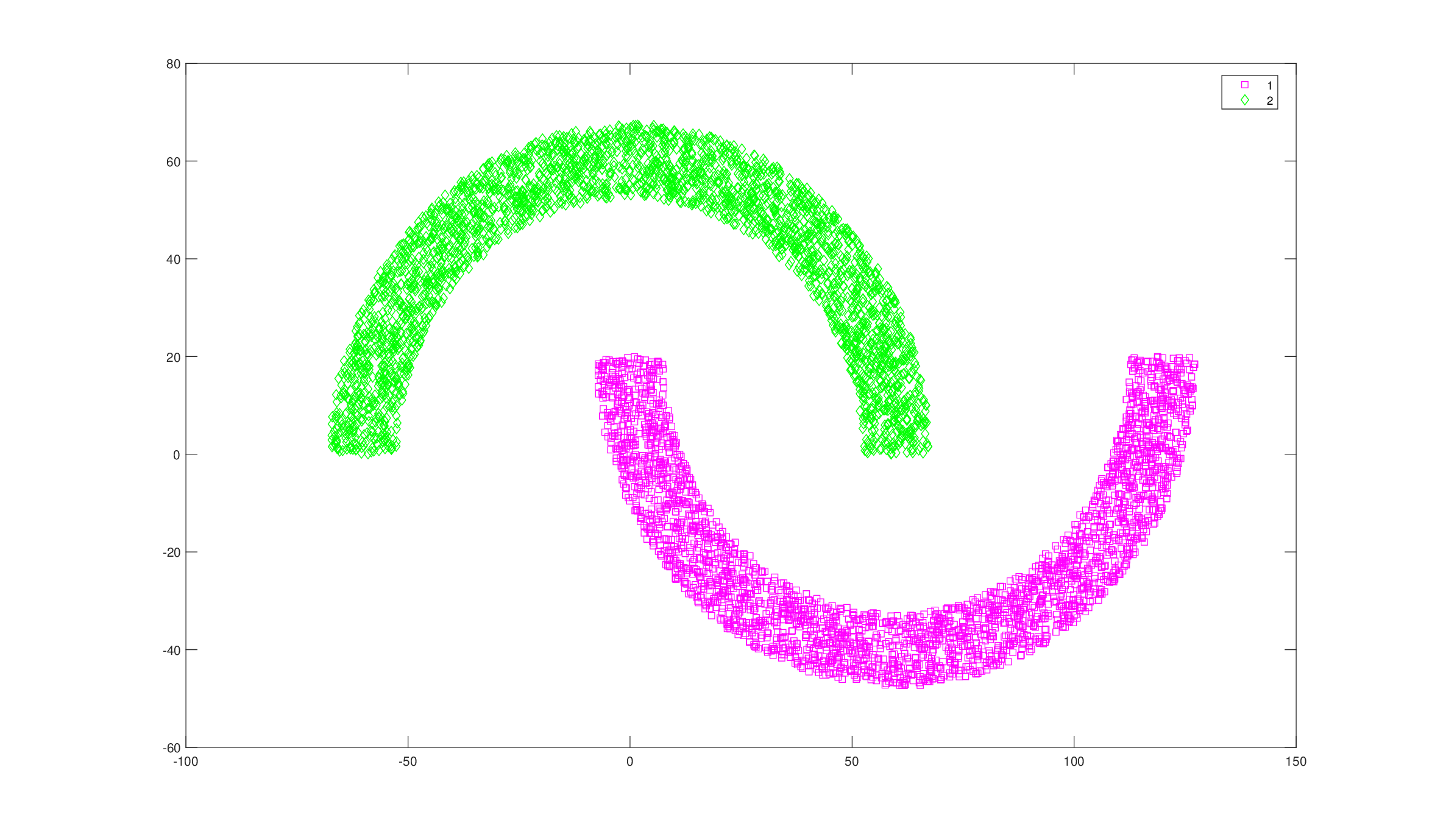} } 
\caption{Clustering results of $k$-means on the original and spectral-embedded feature spaces.} \label{fig:nonconvexshape} 
\end{figure}

According to spectral graph theory \cite{chung1997spectral}, the spectrum of the graph Laplacian contains valuable information about the clustering structure of the graph, which is essential for identifying clusters. For instance, in artificially generated datasets such as two-moons and two-circles, the clustering results of $k$-means applied directly in the original feature space differ significantly from those in the spectrally-embedded feature space, as shown in Fig.~\ref{fig:nonconvexshape}. In the original space, the traditional $k$-means algorithm is unable to correctly identify the clusters due to its reliance on Euclidean distance alone. However, by incorporating spectral information into the feature space, the clustering becomes accurate and aligns with the true underlying structure.

\section{Methods}

\subsection{Algorithmic Framework}

To leverage spectral graph theory for analyzing the key structural information of a dataset, we first construct a graph from the data. For example, a k-nearest neighbor (kNN) graph can be used to provide a basic representation of the dataset. The graph is then embedded into an $r$-dimensional space by utilizing the first $r$ eigenvectors of its Laplacian matrix, a process widely known as spectral graph embedding~\cite{belkin2003laplacian}. Following this, we compute the similarity between data points in the spectral-embedded space, which captures the structural information of the dataset. Next, the samples are divided into subsets such that samples within the same subset exhibit high spectral correlation. The mean of the feature vectors within each subset is then calculated to create a spectrally representative pseudo-sample. These pseudo-samples collectively form a compressed representation of the dataset. If further compression is required, the pseudo-samples can be iteratively compressed again.

The spectral similarity between two points, $u$ and $v$, is defined as follows~\cite{livne2012lean}:

\begin{equation}
s_{uv} := \frac{\left|(\mathbf{X}_u, \mathbf{X}_v)\right|^2}{(\mathbf{X}_u, \mathbf{X}_u)(\mathbf{X}_v, \mathbf{X}_v)}, \quad (\mathbf{X}_u,\mathbf{X}_v) := \sum_{k=1}^{K} \mathbf{x}_u^{(k)} \cdot \mathbf{x}_v^{(k)}.
\end{equation}

In this formulation, $\mathbf{x}_u$ and $\mathbf{x}_v$ represent the feature vectors of points $u$ and $v$ in the spectral embedding space. From a statistical perspective, $s_{uv}$ quantifies the proportion of variance accounted for by the linear regression of $\mathbf{x}_u$ on $\mathbf{x}_v$ in a spectral context. Higher values of $s_{uv}$ signify a stronger spectral association between the two points.

The DBSCAN algorithm is then applied to the compressed data points, grouping them into distinct clusters. Subsequently, each original data point inherits the cluster membership of its corresponding compressed point, thereby finalizing the clustering process.

\section{Experiments}



\begin{table*}[!htbp] 
\begin{center}
\resizebox{\textwidth}{!}{ 
\begin{tabular}{|c|c|c|c|c|c|c|c|}
\hline
\textbf{Data Set} & \textbf{Orig} & \textbf{Ours(2X)} & \textbf{Ours(5X)} & \textbf{Ours(10X)}  \\
\hline
Pendigits &62.1831&62.5967&63.0638&68.5615  \\
\hline
USPS     &81.5552&87.7178&80.3936&79.9419
    \\
\hline
MNIST    &68.5257&78.6400&78.4257&79.7214 
  \\
\hline
\end{tabular}}
\end{center}
\caption{Clustering Accuracy}
\label{table:acc_result}
\end{table*}

\begin{figure}[htbp]\centering
\includegraphics[scale=0.4]{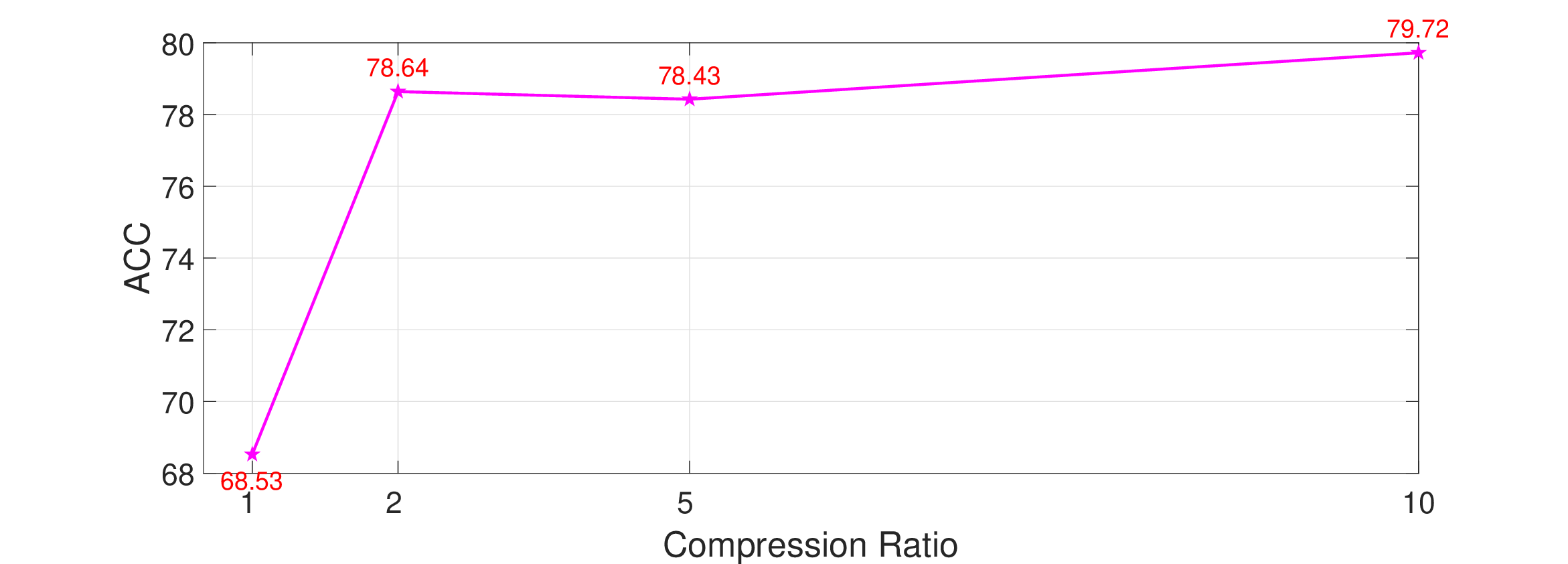}
\caption{Clustering quality VS compression ratio for the MNIST data set.\protect\label{fig:acc_dbscan.eps}}
\end{figure}

\begin{figure}[htbp]\centering
\includegraphics[scale=0.4]{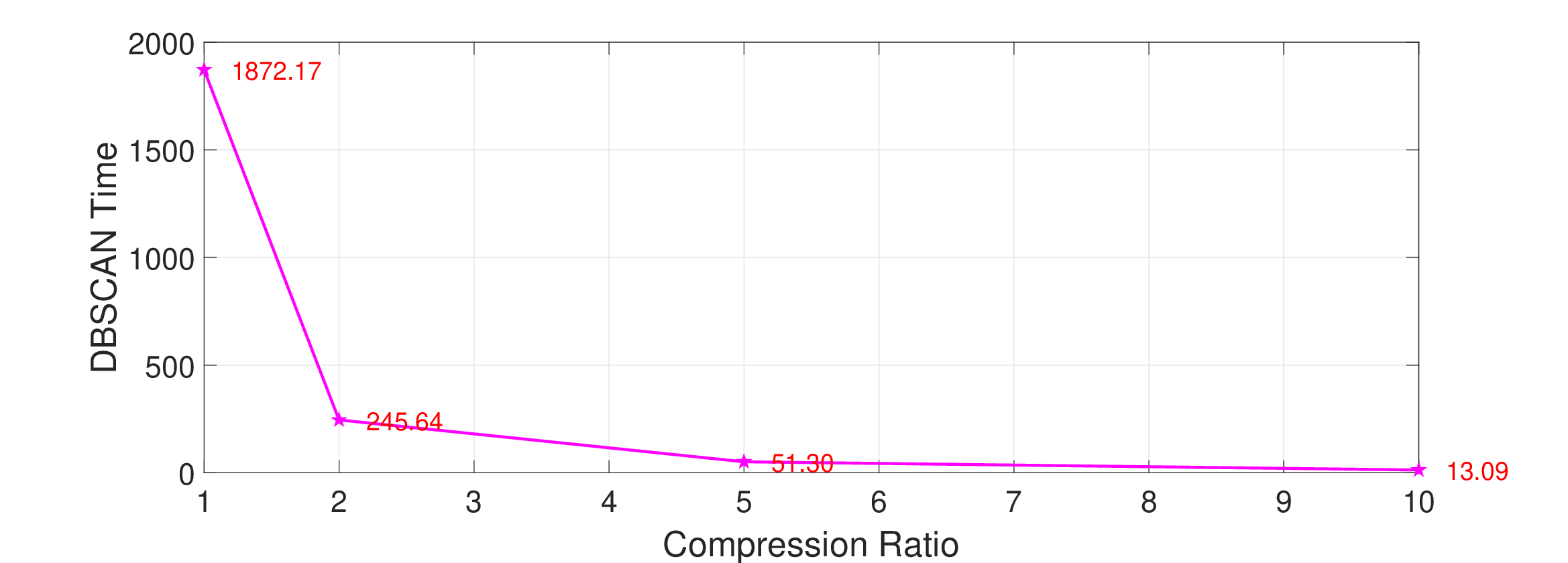}
\caption{DBSCAN time VS compression ratio for the MNIST data set.\protect\label{fig:runtime_dbscan.eps}}
\end{figure}

As illustrated in Fig. \ref{fig:acc_dbscan.eps}, the proposed approach maintains consistently high clustering quality as the compression ratio increases, highlighting its robustness. Furthermore, Fig. \ref{fig:runtime_dbscan.eps} demonstrates that our method achieves significant speed enhancements for DBSCAN algorithm. In addition to these improvements, the method greatly enhances memory and storage efficiency, making it particularly suitable for resource-constrained environments. These advancements pave the way for handling substantially larger datasets on energy-efficient computing platforms, such as FPGAs or handheld devices, emphasizing the transformative potential of the proposed method.

\section{Conclusion}\label{sect:conclusions}
To address the computational challenges of DBSCAN, this study presents a novel spectral data compression approach that creates a significantly smaller data set while preserving the essential spectral properties needed for clustering. Experiments on large-scale, real-world data sets demonstrate that the proposed method can significantly improve the performance of DBSCAN algorithm.

%
%
%
%

\end{document}